\documentclass[11pt]{article}
\pdfoutput=1  

\usepackage[margin=1in]{geometry}
\usepackage{amsmath,amssymb}
\usepackage{booktabs}
\usepackage{graphicx}
\usepackage{xcolor}
\usepackage{hyperref}
\hypersetup{colorlinks=true, linkcolor=blue!60!black, citecolor=blue!60!black, urlcolor=blue!60!black}
\usepackage[numbers]{natbib}

\newcommand{\quality}{\ensuremath{Q}}

\title{Why RAGs Hallucinate: Penalty-Aware Evaluation of\\
Retrieval-Augmented Generation Systems with\\
Knowledge-Gap Canaries}

\author{%
  Alden Do Rosario \and Hussein Younes \and Felipe Pires\\[2pt]
  \normalsize CustomGPT.ai\\
  \normalsize Correspondence: \texttt{alden@customgpt.ai}
}
\date{August 2026}

\begin{document}
\maketitle

\begin{abstract}
Volume-based accuracy rewards retrieval-augmented generation (RAG) systems
for guessing: a system that answers everything outscores one that declines
when its knowledge base cannot support an answer. Building on the
confidence-target analysis of \citet{openai2025hallucinate}, we present a
penalty-aware evaluation framework for \emph{deployed} RAG products,
combining (i) asymmetric scoring (correct $+1$, wrong $-4$, abstain $0$),
(ii) \emph{knowledge-gap canaries}, questions whose answers are verifiably
absent from the knowledge base, so that any answer constitutes ungrounded
generation from parametric memory, and (iii) a failure-attribution pipeline
that separates retrieval, generation, and abstention-policy failures.
Applying the framework to three commercial RAG systems and a no-retrieval
baseline on SimpleQA-Verified (1{,}000 questions $\times$ 3 repeats, graded
blind by a cross-family three-judge panel with 98.9\% unanimity), we find
that accuracy when answering is closely clustered across systems
(97.0--98.0\%), while canary violation rates differ roughly sixfold (16.7\%
vs.\ 98.1\%). The systems are separated less by what they answer correctly
than by whether they answer at all when they should not, and penalty-aware
scoring reorders the volume-based ranking accordingly; OpenAI RAG holds the
top position at every penalty setting from $k{=}1$ to $k{=}9$. All code,
configurations, transcripts, and judge votes are released for independent
audit.
\end{abstract}

\begin{table}[h]
\centering
\caption{Campaign summary: 1{,}000 questions $\times$ 3 repeats, blind
three-judge panel. Scoring: correct $+1$ $\cdot$ wrong $-4$ $\cdot$ abstain
$0$. Best RAG value per column in bold, except the canary column, where the
two lowest rates are not statistically distinguishable
(\S\ref{sec:results}); confidence intervals and significance tests in
\S\ref{sec:results}.}
\label{tab:scoreboard}
\small
\begin{tabular}{lccccc}
\toprule
System & $\quality$ $\uparrow$ & Accuracy $\uparrow$ & Abstention & Acc.$\mid$att.\ $\uparrow$ & Canary viol.\ $\downarrow$\\
\midrule
OpenAI RAG       & $\mathbf{+0.862}$ & $0.948$ & $3.3\%$ & $\mathbf{98.0\%}$ & $22.2\%$\\
Gemini RAG       & $+0.793$ & $\mathbf{0.968}$ & $0.2\%$ & $97.0\%$ & $98.1\%$\\
CustomGPT.ai RAG & $+0.767$ & $0.866$ & $11.1\%$ & $97.5\%$ & $16.7\%$\\
No-RAG Baseline  & $-1.933$ & $0.286$ & $16.0\%$ & $34.0\%$ & exempt\\
\bottomrule
\end{tabular}
\end{table}

\section{Introduction}\label{sec:intro}
Retrieval-augmented generation (RAG) is the dominant strategy for deploying
language models over private knowledge: ground the model in a curated
corpus, and its answers inherit the corpus's authority. Whether a deployed
RAG product actually behaves this way is difficult to determine from
published evaluations. The metrics used to compare such systems (accuracy
over attempted answers, or accuracy over all questions) score an ungrounded
guess and a grounded answer identically when both happen to be right, and
score a confident fabrication and an honest ``I don't know'' identically,
at zero, when the system lacks the answer.
\citet{openai2025hallucinate} formalize the consequence for binary-graded
benchmarks: under such scoring, guessing strictly dominates abstaining, so
models are trained and selected into hallucination by the evaluations
themselves.

This paper applies that analysis to deployed, commercial RAG products,
where the incentive problem is compounded by two features of the product
setting. RAG systems are procured for trustworthiness; the implicit
contract is ``answer from my documents or say you cannot,'' and an
evaluation that cannot observe abstention behavior cannot measure whether
the contract is honored. In addition, the backbone model's parametric
knowledge confounds measurement on public-facts benchmarks: a RAG system
can answer questions its knowledge base does not cover, be scored correct,
and silently substitute world knowledge for grounding. A benchmark that
does not control for this measures the backbone model rather than the
retrieval product.

We address both problems with an evaluation framework built from three
components: asymmetric penalty scoring that makes abstention rational below
an explicit confidence threshold (\S\ref{sec:scoring}); \emph{knowledge-gap
canaries}, questions whose answers verifiably do not exist in the knowledge
base, so that any answer (including a factually correct one) is measurable
parametric leakage (\S\ref{sec:canaries}); and a blind, cross-family,
three-judge grading panel with quantified agreement and a pre-registered
failure-repair policy (\S\ref{sec:judges}). We apply the framework to three
commercial RAG systems and a no-retrieval baseline over 1{,}000
SimpleQA-Verified questions, three full repeats, with every provider call,
judge vote, and scoring decision released for audit.

Empirically, the three RAG systems cluster closely in accuracy when they
choose to answer (97.0--98.0\%), but differ by a factor of more than fifty
in how often they decline (abstention rates of 0.2\% vs.\ 11.1\%) and
roughly sixfold in how often they answer questions their knowledge base
cannot support (canary violation rates of 16.7\% vs.\ 98.1\%). In this
benchmark, the largest observed difference among RAG systems was abstention
behavior rather than accuracy conditional on answering, and volume-based
rankings order systems against it.

\paragraph{Contributions.}
\begin{enumerate}
  \item A penalty-aware scoring framework for deployed RAG systems,
        operationalizing the confidence-threshold analysis of
        \citet{openai2025hallucinate} at the product level
        (\S\ref{sec:method}).
  \item The \emph{canary methodology}: KB-absent questions that convert
        parametric-memory leakage into a directly measurable violation rate
        (\S\ref{sec:canaries}).
  \item A cross-family LLM judge panel with majority voting, agreement
        reporting, and a pre-registered quota-repair policy; we quantify
        same-family judge bias and show the panel's tie-break mechanism
        corrected it (\S\ref{sec:judges}).
  \item A three-repeat, fully audited comparison of three commercial RAG
        products and a no-RAG baseline on SimpleQA-Verified, with
        bootstrap CIs and Bonferroni-corrected significance
        (\S\ref{sec:results}).
\end{enumerate}

\section{Related Work}\label{sec:related}
\paragraph{Hallucination and calibration.}
Hallucination in generative models is broadly surveyed by \citet{ji2023survey};
benchmarks such as TruthfulQA \citep{lin2022truthfulqa} and HaluEval
\citep{li2023halueval} measure susceptibility to imitative falsehoods and
detection of fabricated content. \citet{kadavath2022know} show that models
carry usable self-knowledge about answer correctness, implying abstention is
learnable; \citet{openai2025hallucinate} explain why standard evaluations
suppress it (binary grading makes guessing dominant) and propose
explicit confidence targets in scoring, the analysis our framework
operationalizes for RAG products.

\paragraph{Factuality benchmarks.}
SimpleQA \citep{wei2024simpleqa} established short-form factuality
measurement with adversarially collected, single-answer questions;
SimpleQA-Verified \citep{simpleqaverified2025} re-verified answers and
removed ambiguous items. We use the verified set and extend it with
per-question knowledge-base coverage flags, which is what makes the canary
construction possible.

\paragraph{RAG evaluation.}
RAGAS \citep{es2023ragas}, ARES \citep{saadfalcon2023ares}, and FActScore
\citep{min2023factscore} decompose RAG quality into faithfulness,
relevance, and atomic factual precision. These frameworks evaluate
\emph{pipelines} the evaluator controls and do not model the incentive
structure of answering versus declining; none controls for
parametric-knowledge leakage on questions outside the corpus, the failure
mode our canaries isolate. Complementarily, we evaluate closed commercial
products end-to-end, as procured.

\paragraph{LLM-as-judge.}
LLM judging correlates well with human preference at scale
\citep{zheng2023mtbench} but exhibits self-preference: evaluators favor
their own generations \citep{panickssery2024selfpref}. In our setting the
concern is concrete, since one contestant's backbone shares a family with
any single judge we could pick. This motivates the cross-family panel, the
measured pivotal-vote analysis, and the disclosure regime of
\S\ref{sec:judges}.

\section{Methodology}\label{sec:method}

\subsection{Terminology}\label{sec:terms}
The word ``hallucination'' is used loosely in the literature, and the
distinctions matter for our claims. We use four terms throughout:

\begin{itemize}
  \item \textbf{Factual hallucination}: the answer is wrong.
  \item \textbf{Grounding violation}: the system answers without support
        from its knowledge base, whether or not the answer is factually
        correct.
  \item \textbf{Parametric leakage}: the mechanism behind most grounding
        violations; the answer is produced from the backbone model's
        training knowledge rather than from retrieved content.
  \item \textbf{Canary violation}: the operational measurement; the system
        answers a question whose answer is verified to be absent from the
        knowledge base, so the response is a grounding violation by
        construction.
\end{itemize}

Grounding violations are the mechanism that produces dangerous factual
hallucinations: whenever parametric memory does not happen to contain the
right answer, an ungrounded response becomes a confident fabrication. On
public-facts benchmarks parametric memory is often correct, which is
precisely why grounding violations are invisible to accuracy metrics and
why we measure them directly.

\subsection{Penalty-Aware Scoring}\label{sec:scoring}
Each response receives two scores. The \emph{volume} score is conventional
accuracy: $+1$ if correct, $0$ otherwise. The \emph{quality} score is
asymmetric: correct answers earn $+1$, incorrect answers cost $-k$, and
abstentions score $0$:
\begin{equation}
\quality \;=\; \Pr[\text{correct}] \;-\; k\,\Pr[\text{wrong}],
\qquad k = 4.
\label{eq:quality}
\end{equation}
The penalty ratio implements a confidence target in the sense of
\citet{openai2025hallucinate}. A system that believes its answer is correct
with probability $p$ has expected quality $p - k(1-p)$ from answering versus
$0$ from abstaining; it is rational to answer iff
$p \geq k/(k+1)$. With $k=4$ the indifference point is $p = 0.8$: the
scoring rule rewards exactly the behavior we want from a deployed
knowledge-base assistant: answer when ${\geq}80\%$ confident, otherwise
decline. Volume scoring is the degenerate case $k=0$, under which guessing
is always rational; this is the mechanism by which conventional leaderboards
reward hallucination.

Providers are \emph{not} told about the scoring rule: they respond to bare
questions under a uniform retrieval prompt (\S\ref{sec:setup}), and the
confidence framework is applied post hoc. We therefore measure each
product's \emph{shipped} abstention behavior, not its ability to follow
scoring instructions.

\subsection{Knowledge-Gap Canaries}\label{sec:canaries}
Penalty-aware scoring alone cannot distinguish two failure sources: a wrong
answer may come from faulty retrieval over the knowledge base or from
parametric memory bypassing the knowledge base entirely. We isolate the
second source with \emph{knowledge-gap canaries}: questions from the
evaluation set whose answers are verifiably absent from the knowledge base.
The corpus contains no document from which the answer can be derived;
absence is verified during KB construction and recorded as a per-question
flag. For a KB-grounded system, the correct response to a canary is to
decline.

Any attempted answer to a canary is scored as a violation
($\quality = -k$), including answers that are factually correct. In the
terminology of \S\ref{sec:terms}, a correct canary answer is not a factual
hallucination, but it is a grounding violation produced by parametric
leakage, and it identifies exactly the systems that will fabricate
confidently on corpora where parametric memory has no correct answer to
supply. Volume scores are left untouched by the override, so both metrics
are reported from one run. The no-retrieval baseline is exempt because it
has no knowledge base to be grounded in. Our evaluation set contains 18
canaries; across 3 repeats each provider faces 54 canary trials.

\subsection{Evaluation Pipeline and Blind Protocol}\label{sec:pipeline}
Each provider response passes through three stages. \textbf{(1) Intent
classification:} a lightweight model (gpt-5.4-nano, reasoning disabled)
classifies the response as an \emph{attempt} or an \emph{abstention} using
few-shot criteria; abstentions score $0$ and are not judged. Classifier
failures fall back to conservative phrase heuristics rather than a silent
default. \textbf{(2) Panel grading} (\S\ref{sec:judges}) for attempts.
\textbf{(3) Scoring} per \S\ref{sec:scoring}--\ref{sec:canaries}.

Judging is blind: providers are shown to judges only as anonymous IDs, the
question order is shuffled with a fixed seed, and the identity mapping is
revealed only after all grading completes. Technical failures (API errors,
deterministic empty responses) are excluded from scoring rather than counted
as abstentions, and every exclusion is disclosed.

\subsection{Cross-Family Judge Panel}\label{sec:judges}
A single-family LLM judge cannot be neutral in a study that includes
products backed by its own model family. We therefore grade every attempt
with a three-judge panel spanning the three major model families
(gpt-5.4, Claude Sonnet~5, and Gemini~3 Flash), each grading independently
(binary correct/incorrect against the gold answer, with free-text
reasoning and a confidence estimate), with the majority verdict final.
Every judge shares a model family with one of the systems under test, and
two share a checkpoint: the GPT judge and the OpenAI provider both run
\texttt{gpt-5.4}, and the Gemini judge and the Gemini provider both run
\texttt{gemini-3-flash-preview}. Only the Claude judge (Sonnet~5) differs
from the corresponding backend (the CustomGPT.ai backbone is Sonnet~4.6).
Cross-family majority voting is therefore a mitigation for self-preference,
not a guarantee of independence. For the GPT judge we measure the effect
directly: its vote was pivotal in favor of an OpenAI product in $1/544$
OpenAI gradings ($0.2\%$) in pre-campaign runs, and provider rankings are
unchanged if the GPT judge is excluded entirely. The corresponding
leave-one-out check for the Gemini judge is not reported here; every
per-judge vote is released so that it can be computed independently.

A failed judge cannot stall grading: each judge call carries a hard
wall-clock timeout, and a failed or timed-out judge is recorded as an
invalid vote while the remaining judges' majority decides. A 2--0 majority
cannot be overturned by the missing vote. Ties, which can arise only when a
judge fails, fall to the designated primary judge. A held-out consistency
validation re-grades sampled responses three times per provider to measure
judge determinism.

\paragraph{Pre-registered quota repair.} Third-party judge APIs impose
per-day quotas that a large campaign can exhaust mid-run. Our pre-registered
policy: the run completes on the remaining judges' majorities, and the
missing votes are re-cast on the following quota day \emph{from the verbatim
stored judge prompts}, with provider answers never regenerated, after
which panels are re-tallied with the identical rules. In the campaign this
affected 815 of 11{,}080 gradings (all in one repeat); backfilling flipped
12 final grades, exactly the 12 that had been decided by a two-judge
tie-break, 10 of them \emph{toward} the vendor-affiliated product. The
completed panels thus corrected a residual bias in the degraded mode rather
than introducing one, and post-repair scores for the affected repeat fall
within the range of the unaffected repeats.

\subsection{Abstention Forensics}\label{sec:forensics}
Aggregate scores say \emph{that} a system abstained or erred, not
\emph{why}. For every penalized response the pipeline captures the retrieval
evidence available to the provider (returned citations or grounding chunks
where the API exposes them) and attributes the failure: answer absent from
retrieved context (retrieval failure), answer present but not used
(generation failure), or judging artifact. In pilot runs this attribution
showed that one third of the vendor product's failures were abstentions
despite the answer being present in retrieved context. This over-caution
mode is invisible to score-only evaluation, and the observation motivated
reporting abstention behavior as a first-class result.

\subsection{Audit Trail}\label{sec:audit}
Every experimental claim is reconstructible from append-only logs written
during the run: all provider requests (with latency, token usage, and cost),
all judge votes (with prompts, reasoning, and panel tallies), all intent
classifications, consistency-validation records, and run configuration. The
released statistics are computed by an independent script that rebuilds
per-response scores from these logs and is validated to reproduce the
pipeline's own aggregates exactly before computing campaign-level
statistics.

\section{Experimental Setup}\label{sec:setup}

\subsection{Dataset and Knowledge Base}
We evaluate on SimpleQA-Verified \citep{simpleqaverified2025}, 1{,}000
short-form factual questions curated from SimpleQA \citep{wei2024simpleqa}
for unambiguous gold answers. The knowledge base was constructed by
collecting one web-sourced document per question, giving a corpus of
1{,}000 documents. This corpus was uploaded, unchanged, to each vendor's
retrieval store (CustomGPT.ai project, OpenAI vector store, and Google File
Search store), so all systems retrieve over identical content.
After construction, every question was checked against the corpus for
answer coverage. For 982 questions the collected document contains the gold
answer. For the remaining 18 questions the collected document does not
(the source page failed to state the target fact), so their answers are
absent from the corpus. Rather than repairing these gaps, we flag the 18
questions as the canary set of \S\ref{sec:canaries}; the flag is consumed
by the scoring pipeline.

\subsection{Systems Under Test}
We evaluate the RAG products \emph{as shipped}: each vendor's own chunking,
embedding, retrieval, and generation stack, configured only through public
APIs. Table~\ref{tab:roster} lists the system roster. Each provider runs
its current generally available, flagship-class backbone; the roster was
frozen in July 2026 and the campaign executed against it in August 2026.
Reasoning is disabled where the API supports it, or set to the provider's
lowest available setting otherwise.

\begin{table}[t]
\centering
\caption{System roster (frozen July 2026; campaign executed August 2026).
All backbones are pinned, documented, and verified at run start; the
CustomGPT.ai backend pin is re-checked against the vendor's settings API
before every run and the run aborts on mismatch.}
\label{tab:roster}
\begin{tabular}{llll}
\toprule
System & Backbone & Reasoning setting & Retrieval\\
\midrule
CustomGPT.ai RAG & Claude Sonnet 4.6 & non-thinking variant & vendor-managed\\
OpenAI RAG & gpt-5.4 & \texttt{reasoning: none} & file\_search / vector store\\
Gemini RAG & gemini-3-flash-preview & \texttt{thinking: MINIMAL} & File Search\\
No-RAG Baseline & gpt-5.4 & \texttt{reasoning: none} & none\\
\bottomrule
\end{tabular}
\end{table}

\paragraph{Parity measures.} All systems receive the same questions in the
same fixed-seed order, a fresh session per question (no conversational
carry-over), and temperature $0$ where the API accepts it. Latency and cost
are recorded per request. The prompt variant used throughout the campaign is
the repository's \texttt{standard} variant; a stricter variant exists in the
released code and was not used, so the abstention behavior we report is the
behavior these products exhibit under a neutral instruction rather than under
an instruction to refuse.

\paragraph{Prompt asymmetry, disclosed.} Prompt delivery is \emph{not}
symmetric across systems, and the asymmetry falls on the vendor-affiliated
product. OpenAI RAG and Gemini RAG each receive an identical system message
instructing grounding in the knowledge base together with an explicit decline
phrasing; the no-retrieval baseline receives the analogous no-KB prompt. The
CustomGPT.ai platform does not accept a per-request system message: its
instruction is a project-level persona, and the decline clause was therefore
never propagated to it. We discovered this after the campaign, from the audit
logs. The direction is conservative with respect to our own product: the only
RAG system that was \emph{not} instructed to decline is the one that declined
most often ($11.1\%$, against $3.3\%$ and $0.2\%$) and violated the fewest
canaries. Readers should nonetheless read CustomGPT.ai's abstention rate as
shipped-product behavior and the other two as partly prompt-conditioned. All
three verbatim prompts are recorded in the released run metadata.

Two further deviations from strict symmetry are disclosed. Google's Pro-tier
model was unusable at our API tier (250 requests/day project quota) and
exhibited no latency-class advantage under load; the Gemini system therefore
runs a Flash model, and cross-vendor price asymmetry ($\$0.50$--$\$3$ vs.\
$\$2.50$--$\$15$ per MTok) is noted where relevant. Output budgets are capped
at 1{,}024 tokens for every system except Gemini, whose call is uncapped; the
longest Gemini completion in the campaign is 479 tokens, so no response
approaches the cap and the asymmetry is immaterial.

\subsection{Judge and Classifier Configuration}
All three judges run deterministic configurations (temperature $0$ or the
API's deterministic default; fixed seed where supported; JSON-schema or
JSON-mode output) with reasoning disabled, and are given only the question,
gold answer, and anonymized response. The intent classifier
(\S\ref{sec:pipeline}) runs gpt-5.4-nano with reasoning disabled. Exact
request parameters for every model are recorded in the audit logs and
released.

\subsection{Campaign Protocol}\label{sec:protocol}
Because the definitive run is expensive and must not be selectively
re-run, we validated the pipeline first on a 107-question hard set mined
from all prior pilot runs: questions any system had answered incorrectly or
wrongly abstained on in at least two runs, plus all canaries. This
validation set was executed end-to-end on the frozen configuration until
the full evidence chain was clean, then executed twice to verify score
stability (observed score drift $\leq 0.16$ on the hard set, with no rank
changes). The validation runs surfaced five pipeline defects that were
fixed before the campaign, including a silently failing abstention
classifier whose fallback would have scored every abstention as a wrong
answer.

The definitive campaign is $1{,}000$ questions $\times$ 4 systems $\times$
3 repeats under a single-launch protocol: the configuration is frozen
before launch, repeats are part of the design (variance estimation), no
result-conditioned re-running is permitted, and every anomaly is logged and
disclosed rather than patched. Provider answering and panel judging
consumed about 16{,}000 provider calls and 33{,}000 judge votes; total API
cost was under \$900.

\section{Results}\label{sec:results}

\begin{table}[t]
\centering
\caption{Main results: 1{,}000 questions $\times$ 3 repeats. Penalty-aware
score $\quality =\Pr[\text{correct}]-4\Pr[\text{wrong}]$; brackets are
bootstrap 95\% CIs (10{,}000 resamples over questions). Canary rate is the
fraction of 54 canary trials (18 questions $\times$ 3 repeats) answered
rather than declined, with Wilson 95\% intervals. \quality{} is not
recoverable from the Accuracy and Abstain columns alone: the canary override
of \S\ref{sec:canaries} additionally converts each factually correct canary
answer from $+1$ to $-4$. Those conversions number 6 of 12 attempted canary
trials for OpenAI RAG, 33 of 53 for Gemini RAG, and 6 of 9 for CustomGPT.ai
RAG; applying them to $\Pr[\text{correct}]-4\Pr[\text{wrong}]$ reproduces
the reported \quality{} values. The baseline is canary-exempt and needs no
adjustment.}
\label{tab:main}
\begin{tabular}{lccccc}
\toprule
System & $\quality$ [95\% CI] & Accuracy & Abstain & Acc.$\mid$att. & Canary viol. [95\% CI]\\
\midrule
OpenAI RAG   & $+0.862$ [$+0.816,+0.903$] & $0.948$ & $3.3\%$  & $98.0\%$ & $22.2\%$ [$13.2,34.9$]\\
Gemini RAG   & $+0.793$ [$+0.730,+0.853$] & $\mathbf{0.968}$ & $0.2\%$ & $97.0\%$ & $98.1\%$ [$90.2,99.7$]\\
CustomGPT.ai RAG & $+0.767$ [$+0.717,+0.812$] & $0.866$ & $11.1\%$ & $97.5\%$ & $16.7\%$ [$9.0,28.7$]\\
No-RAG Baseline & $-1.933$ [$-2.066,-1.797$] & $0.286$ & $16.0\%$ & $34.0\%$ & exempt\\
\bottomrule
\end{tabular}
\end{table}

\subsection{Overall Comparison}
Table~\ref{tab:main} presents the main results. In this benchmark, the
largest effect by far is retrieval itself: every RAG system scores about
$2.7$ points above the no-retrieval baseline on the penalty-aware metric
($p = 10^{-4}$, paired bootstrap), with wrong-answer rates falling from
about $56\%$ to $2$--$3\%$. Among RAG systems, OpenAI RAG has the highest
$\quality$; its difference over CustomGPT.ai is significant after
Bonferroni correction, its difference over Gemini narrowly misses the
corrected threshold, and CustomGPT.ai and Gemini are not significantly
different from one another (Table~\ref{tab:pairwise}).

\begin{table}[t]
\centering
\caption{Pairwise quality differences (paired by question, averaged over
repeats; bootstrap with 10{,}000 resamples; Bonferroni-corrected
$\alpha = 0.05/6 = 0.0083$).}
\label{tab:pairwise}
\begin{tabular}{lccc}
\toprule
Comparison & $\Delta\quality$ [95\% CI] & $p$ & Significant\\
\midrule
OpenAI RAG vs.\ CustomGPT.ai & $+0.095$ [$+0.048, +0.142$] & $10^{-4}$ & yes\\
OpenAI RAG vs.\ Gemini    & $+0.069$ [$+0.015, +0.125$] & $0.0084$ & no (marginal)\\
CustomGPT.ai vs.\ Gemini     & $-0.026$ [$-0.090, +0.037$] & $0.40$ & no\\
Each RAG system vs.\ Baseline & ${\approx}+2.7$ & $10^{-4}$ & yes\\
\bottomrule
\end{tabular}
\end{table}

\subsection{Penalty Sensitivity}\label{sec:ksens}
The penalty $k=4$ encodes one particular confidence target, so we recompute
the full ranking under $k \in \{1, 2, 4, 9\}$, corresponding to confidence
thresholds of 50\%, 67\%, 80\%, and 90\% (Table~\ref{tab:ksens}). OpenAI
RAG holds the top position at every penalty level. Gemini's position
degrades monotonically as the penalty grows, from second at $k{=}1$ to last
among RAG systems at $k{=}9$, while CustomGPT.ai's improves; the two cross
between $k{=}4$ and $k{=}9$. The direction of the effect is the point: the
stricter the confidence requirement, the more a never-abstain policy costs,
and the conclusions of this paper do not depend on the specific choice
$k{=}4$.

\begin{table}[t]
\centering
\caption{Ranking by penalty-aware score under different penalties (mean
$\quality$ per system; baseline omitted, it is last at every $k$).}
\label{tab:ksens}
\begin{tabular}{lcccc}
\toprule
System & $k{=}1$ (50\%) & $k{=}2$ (67\%) & $k{=}4$ (80\%) & $k{=}9$ (90\%)\\
\midrule
OpenAI RAG & $\mathbf{+0.925}$ & $\mathbf{+0.904}$ & $\mathbf{+0.862}$ & $\mathbf{+0.757}$\\
Gemini RAG & $+0.916$ & $+0.875$ & $+0.793$ & $+0.588$\\
CustomGPT.ai RAG & $+0.840$ & $+0.815$ & $+0.767$ & $+0.645$\\
\bottomrule
\end{tabular}
\end{table}

\subsection{Accuracy and Penalty-Aware Score Disagree}
\begin{figure}[t]
\centering
\includegraphics[width=0.72\linewidth]{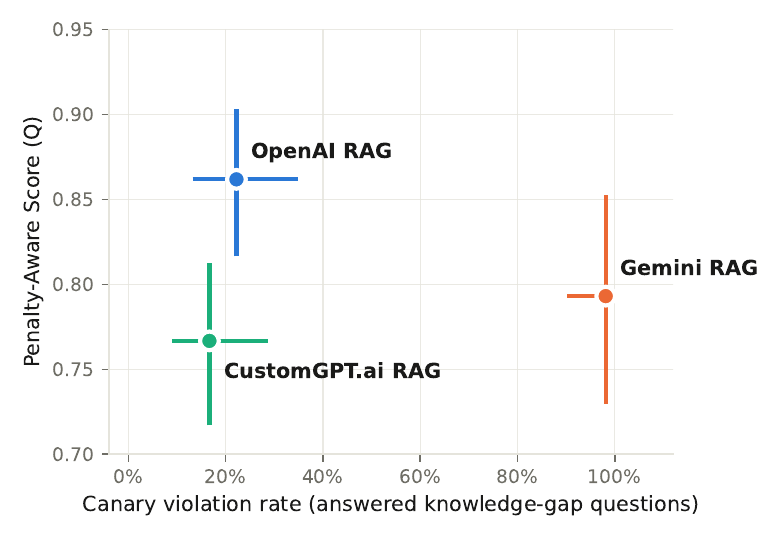}
\caption{Quality score versus canary violation rate for the three RAG
systems (error bars: bootstrap 95\% CI on quality; Wilson 95\% interval on
violation rate). The no-RAG baseline (quality $-1.93$) is canary-exempt and
omitted. Gemini buys the highest answer volume by answering KB-absent
questions from parametric memory; the penalty-aware metric prices that
behavior in.}
\label{fig:inversion}
\end{figure}

Conventional accuracy and the penalty-aware score produce different
orderings, and the canaries explain why (Figure~\ref{fig:inversion}).
Gemini attains the highest accuracy (0.968) by answering essentially
everything: it abstained on 2 of 3{,}000 answers and attempted $98.1\%$ of
canary trials, questions whose answers our knowledge-base construction check
found to be absent, sourcing those answers from parametric memory.
CustomGPT.ai and OpenAI RAG attempted only $16.7\%$ and $22.2\%$ of canary
trials respectively; the Wilson intervals of both are disjoint from Gemini's.
The two are not, however, distinguishable from each other: at the question
level the difference is 3 canaries versus 4 out of 18, which no test
separates (Fisher exact $p = 1.00$). The canary result that this study
supports is Gemini versus the other two systems, not an ordering within that
pair. On a
public-facts benchmark, parametric answers are often factually correct, so
accuracy metrics reward the behavior. On private corpora, where parametric
memory has nothing correct to supply, the same policy would produce
fabricated answers; we return to this implication, which our data do not
directly test, in \S\ref{sec:discussion}. The canary violation rate
separates the systems more sharply than any other measurement in the
study, and penalty-aware scoring translates it into the reordering visible
in Table~\ref{tab:main}.

\subsection{Abstention Behavior Separates the Systems}
Accuracy given an attempt is closely clustered across the three RAG
systems: $98.0\%$ [Wilson 95\% CI $97.5$--$98.5$] for OpenAI RAG, $97.5\%$
[$96.8$--$98.0$] for CustomGPT.ai, and $97.0\%$ [$96.3$--$97.5$] for
Gemini. Abstention rates, by contrast, span a fiftyfold range: CustomGPT.ai
declines $11.1\%$ of questions, OpenAI RAG $3.3\%$, and Gemini $0.2\%$. In
this benchmark, the largest observed difference among the RAG systems was
abstention behavior rather than accuracy conditional on answering, and
evaluations that do not price abstention cannot observe it.

\subsection{Stability Across Repeats}
Rankings are unchanged in every repeat on both metrics. Per-provider quality
drifts by at most $0.034$ between repeats (OpenAI RAG: $0.017$ across three
repeats), canary violation counts vary by at most one trial, and abstention
counts by ${\leq}4$ per 1{,}000. Provider non-determinism exists (roughly
$1$--$3\%$ of per-question outcomes change between repeats) but it is an
order of magnitude smaller than the inter-provider differences we report.

\subsection{Judge Reliability}
Across 11{,}080 panel gradings the three judges were unanimous on $98.9\%$;
no grading required a tie-break in the completed panels. Each judge's
disagreement rate with the final verdict is at most $0.6\%$ (GPT $0.6\%$,
Gemini $0.4\%$, Claude $0.0\%$), and the held-out consistency validation
found no evidence of judge non-determinism on re-graded samples.

\subsection{Latency and Cost}
All three RAG systems answer in the same latency class under identical
parallel load (median $7$--$12$s; the no-retrieval baseline's $3.0$s shows
retrieval dominates response time), while per-query cost spans two orders
of magnitude (Table~\ref{tab:latcost}).

\begin{table}[t]
\centering
\caption{Per-query latency and cost, measured across all campaign requests
(3{,}000 per system).}
\label{tab:latcost}
\begin{tabular}{lccc}
\toprule
System & Median (s) & p90 (s) & Cost/query\\
\midrule
OpenAI RAG & 9.2 & 12.7 & \$0.047\\
Gemini RAG & 12.4 & 14.8 & \$0.0018\\
CustomGPT.ai RAG & 7.0 & 19.1 & \$0.100 (flat)\\
No-RAG Baseline & 3.0 & 4.7 & \$0.0006\\
\bottomrule
\end{tabular}
\end{table}

\section{Why RAG Systems Hallucinate}\label{sec:why}
The campaign's audit data let us attribute failures to mechanisms, not just
count them. For each response we join the final grade and abstention status
with the retrieval-evidence signal the provider exposes: CustomGPT.ai
returns per-answer citations and Gemini reports whether File Search
grounding chunks were attached. OpenAI's response metadata carries no
comparable public field, so OpenAI RAG appears only in the aggregate
results. Evidence presence is a coarse signal: it records that retrieval
returned sources, not that the sources contain the answer. A chunk-level
check of that stronger property was performed in pilot runs and is
reported below.

\paragraph{Grounding violations come from a policy, not a defect.}
The canary trials measure what a system does when its knowledge base has
nothing to offer. Gemini answered 53 of 54 such trials; CustomGPT.ai
answered 9 and OpenAI RAG 12. The same contrast appears outside the canary
set. Of CustomGPT.ai's 289 non-canary abstentions across the campaign, 258
($89\%$) occurred when its retrieval returned no citations: when retrieval
comes back empty, the system declines. Gemini's grounding metadata reported
attached chunks on effectively every request, and the system declined 4
times in 3{,}000 questions: whatever retrieval returns, it answers. These
are two coherent but opposite policies for the weak-retrieval regime, and
the canary violation rate is the visible consequence. Hallucination risk
in these products is not primarily a generation defect; it is the
configured answer-versus-decline policy applied when grounding is thin.

\paragraph{Wrong answers mostly occur with retrieval evidence attached.}
Non-canary wrong answers are rare for the RAG systems (64 for CustomGPT.ai
and 70 for Gemini across 3{,}000 answers each). Nearly all of them carried
retrieval evidence: $98\%$ of CustomGPT.ai's wrong answers cited sources
and $100\%$ of Gemini's reported grounding chunks. The dominant wrong-answer
mode is therefore not answering from nothing; it is retrieval surfacing
content that does not settle the question, with generation answering
anyway. The pilot-run chunk-level analysis sharpens this: in one third of
the vendor system's pilot failures the gold answer was present in the
retrieved context and the system nonetheless abstained or answered
incorrectly, so generation-side failures occur in both directions
(over-caution and misuse).

\paragraph{A failure taxonomy.}
Combining the mechanisms, ordered by measured frequency: (1) parametric
leakage on knowledge gaps, the largest and most system-dependent source
(0--98\% violation rates by policy); (2) retrieval-supported wrong answers,
where sources are returned but the answer is still wrong (about 2\% of
answers for every RAG system); (3) over-caution, declining despite usable
evidence (11\% of CustomGPT.ai's abstentions carried citations); and (4)
retrieval gaps handled correctly by abstention, which cost accuracy but not
trust. Penalty-aware scoring prices (1) and (2) while leaving (4)
unpunished, which is what the conventional-accuracy metric gets backwards.

\section{Discussion}\label{sec:discussion}
\paragraph{Implications for buyers.}
A procurement decision made on conventional accuracy would select the
system with the highest rate of ungrounded answers. On public facts this is
survivable because parametric memory is often right. We hypothesize that on
the private corpora RAG is typically deployed for (contracts, medical
records, internal documentation) the same never-decline policy would
produce confident fabrication, because parametric memory has no correct
answer to supply; our benchmark does not test private corpora, and
verifying this extrapolation is important future work. What the data do
establish is that abstention and canary behavior separate current products
far more than answer accuracy does, so evaluations of RAG products should
report both as first-class metrics.

\paragraph{Implications for benchmark design.}
Any evaluation that controls its corpus can flag questions whose answers
the corpus lacks and thereby convert parametric leakage into a measurable
rate. Penalty-aware scoring requires no provider cooperation; we applied it
post hoc to shipped products. Public leaderboards could adopt both
components without changing how systems are queried, and the differences we
measure (closely clustered accuracy-given-attempt, fiftyfold spreads in
abstention) suggest doing so would reorder existing RAG rankings.

\paragraph{Implications for LLM judging.}
The panel data give a quantified caution for single-judge evaluation.
Judges from different families agreed on $98.9\%$ of gradings, but the
residual disagreements were not random: in the degraded two-judge mode,
tie-breaks systematically disfavored one vendor's answers, and completing
the panels reversed 12 of 12 such decisions. Cross-family majority voting
with agreement reporting is a low-cost defense, and its value concentrates
on exactly the contested gradings.

\section{Limitations and Conflict of Interest}\label{sec:limitations}
\paragraph{Conflict of interest.}
The authors are affiliated with CustomGPT.ai, the vendor of one evaluated
system. We mitigate rather than deny the conflict: the study is
methodology-first; judging is blind and cross-family with agreement and
pivotal-vote impact reported; every provider request, judge vote,
classification, and scoring decision is released for independent
recomputation; and the affiliated product does not lead the headline
penalty-aware metric. Readers can re-derive every number from the released
logs.

\paragraph{Access parity.}
As the vendor, we hold administrative access to the CustomGPT.ai platform
that we do not hold for competitors. In this study that access was used
for exactly one action: pinning the backend model via the vendor's public
settings API, an endpoint available to any paying account, with the pin
re-verified before every run. All queries to all systems, including
CustomGPT.ai, went through the same public query APIs used by any
customer; no private endpoints, internal configuration, index tuning, or
non-public parameters were used for any system. The knowledge base
uploaded to CustomGPT.ai is byte-identical to the corpus uploaded to the
other vendors' stores.

\paragraph{Reproducibility.}
Code, run configurations, per-request audit logs (prompts, raw responses,
judge votes, classifications, exclusions), the statistics scripts that
regenerate every table and figure, and the document corpus manifest are
public at
\url{https://github.com/adorosario/why-rags-hallucinate}.

\paragraph{Limitations.}
(1) One dataset, English-only, short-form factual questions; behavior on
multi-hop or long-form tasks may differ. (2) The Gemini system runs
Google's flagship Flash model rather than Pro, a constraint imposed by API
quota (250 requests/day at our tier) documented in \S\ref{sec:setup}; Pro
showed no latency-class advantage under load, but we cannot rule out
quality differences. (3) One judge shares its model with the OpenAI
contestant; its measured pivotal impact is $0.2\%$ of that vendor's
gradings and rankings are unchanged when it is excluded. (4) One question
was excluded for one provider (deterministic empty response, disclosed;
$n{=}999$ for CustomGPT.ai). (5) 815 judge votes in one repeat were re-cast
one day late under the pre-registered quota-repair policy
(\S\ref{sec:judges}). (6) The canary count is modest ($18$ questions, $54$ trials per system).
The canaries were not designed in advance; they are the questions whose
collected source documents turned out not to contain the gold answer,
discovered during corpus verification. We kept them rather than repairing
the corpus because they measure exactly the behavior of interest. The
resulting intervals are wide, but the central contrast ($16.7\%$ vs.\
$98.1\%$) is far larger than the interval widths; a deliberately
constructed larger canary set is a natural extension. (7) Commercial
systems are moving targets: results are a snapshot of pinned, documented
configurations (roster frozen July 2026, campaign executed August 2026).

\section{Conclusion}\label{sec:conclusion}
We presented a penalty-aware evaluation framework for deployed RAG systems
built on knowledge-gap canaries and cross-family judging, and applied it in
a fully audited three-repeat campaign across three commercial products and
a no-retrieval baseline. Retrieval produced the largest effect in the
study. Among the RAG systems, accuracy when answering was closely
clustered, and the measured differences concentrated in the policy
governing when not to answer: canary violation rates ranged from $16.7\%$
to $98.1\%$ across products that conventional accuracy ranks in the
opposite order, and the reordering persists for every penalty setting we
tested. Evaluations that do not price abstention are not neutral; they
reward the behavior RAG is deployed to prevent. The framework, code, and
complete audit trail are released so this measurement can be repeated,
contested, and extended.

\bibliographystyle{plainnat}
\bibliography{references}

\end{document}